\pdfoutput=1

\documentclass[pageno]{jpaper}

\usepackage{algorithm}
\usepackage[noend]{algpseudocode}
\usepackage{amsmath}
\usepackage{amssymb}
\usepackage{amsthm}
\usepackage{booktabs} 
\usepackage{dsfont}
\usepackage{float}
\usepackage{listings}
\usepackage{multirow}
\usepackage{perpage} 
\usepackage{pifont}
\usepackage{subfig}
\usepackage[normalem]{ulem}
\usepackage[sort, numbers]{natbib}

\MakePerPage{footnote} 

\usepackage{soul}
\usepackage[normalem]{ulem}
\usepackage{xspace}

\newcommand{\originalgrumbler}[2]{\begin{quote}\textcolor{blue}{\sl{\bf #1 says:} #2}\end{quote}}
\newcommand{\grumbler}[2]{\originalgrumbler{#1}{#2}}
\newcommand{\minjia}[1]{\grumbler{Minjia}{#1}}

\newcommand\later[1]{\begin{quote}\textcolor{green}{\textbackslash \textbf{later\{}} #1 \textcolor{green}{\}}\end{quote}}
\newcommand\notes[1]{\begin{quote}\textcolor{green}{\textbackslash \textbf{notes\{}} #1 \textcolor{green}{\}}\end{quote}}

\newcommand{\sfsmaller}{}

\newcommand{\code}[1]{\textsf{\sfsmaller#1}}

\newcommand{\eg}{e.g.\xspace}
\newcommand{\ie}{i.e.\xspace}

\theoremstyle{definition}
\newtheorem{defn}{Definition}[section]

\newcommand{\cmark}{\ding{51}}%
\newcommand{\xmark}{\ding{55}}%

\newcommand{\ANNeX}{Zoom\xspace}

\newcommand{\IVFPQ}{IVFPQ\xspace}
\newcommand{\cells}{clusters\xspace}
\newcommand{\cell}{cluster\xspace}

\newcommand{\Ncluster}{$N_{cluster}$\xspace}
\newcommand{\Ccluster}{$S_{centroid}$\xspace}

\newcommand{\Npq}{$N_{pq}$\xspace}

\newcommand{\SIFT}{SIFT1M\xspace}
\newcommand{\Deep}{Deep10M\xspace}

\newcommand{\peakperf}{$1.69$Tflops\xspace}

\renewcommand{\grumbler}[2]{}
\renewcommand{\notes}[1]{}
\renewcommand{\later}[1]{}

\newcommand{\outline}[1]{\grumbler{outline}{#1}}
\renewcommand{\outline}[1]{}

\newtoggle{tech-report}
\togglefalse{tech-report}
\iftoggle{tech-report}{%
}{

}

\setlist{nosep} 
\usepackage{titlesec}
\usepackage{setspace}

\newlength{\sectionbelowskip}
\newlength{\sectionaboveskip}
\newlength{\subsectionbelowskip}
\newlength{\subsectionaboveskip}
\newlength{\subsubsectionbelowskip}
\newlength{\subsubsectionaboveskip}
\newlength{\paragraphaboveskip}

\newcommand{\setvspace}[2]{%
  #1 = #2
  \advance #1 by -1\parskip}

\setvspace{\sectionaboveskip}{3pt}
\setvspace{\subsectionaboveskip}{3pt}
\setvspace{\subsubsectionaboveskip}{2pt}
\setvspace{\paragraphaboveskip}{2pt}

\titlespacing*{\section}{0pt}{\sectionaboveskip}{\sectionbelowskip}
\titlespacing*{\subsection}{0pt}{\subsectionaboveskip}{\subsectionbelowskip}
\titlespacing*{\subsubsection}{0pt}{\subsectionaboveskip}{\subsectionbelowskip}
\titlespacing*{\paragraph}{0pt}{\paragraphaboveskip}{*}

\titleformat{\subsubsection}
  {\normalfont\normalsize\bfseries}{\thesubsubsection}{1em}{}
\titlespacing*{\subsubsection}{0pt}{\subsubsectionaboveskip}{\subsubsectionbelowskip}

\makeatletter 
\def\thm@space@setup{%
  \thm@preskip=3pt
  \thm@postskip=\thm@preskip 
}
\makeatother



\begin{document}

\title{Zoom: SSD-based Vector Search for Optimizing Accuracy, Latency and Memory\vspace{-1em}}
\author{%
Minjia Zhang
\qquad
Yuxiong He
\qquad
\\Microsoft 
\\{\textit {\{minjiaz,yuxhe\}@microsoft.com}}\vspace{-2em}
}


%
%



\date{}

\maketitle

\begin{abstract}
With the advancement of machine learning and deep learning, vector search becomes instrumental to many information retrieval systems, to search and find best matches to user queries based on their semantic similarities.
These online services require the search architecture to be both effective with high accuracy and efficient with low latency and memory footprint, which existing work fails to offer.
We develop, \ANNeX, a new vector search solution that collaboratively optimizes accuracy, latency and memory based on a multiview approach.  (1) A "preview" step generates a small set of good candidates, leveraging compressed vectors in memory for reduced footprint and fast lookup.  (2) A "fullview" step on SSDs reranks those candidates with their full-length vector, striking high accuracy.
Our evaluation shows that, \ANNeX achieves an order of magnitude improvements on efficiency while attaining equal or higher accuracy, comparing with the state-of-the-art. 
\end{abstract}

\section{Introduction}
\label{sec:intro}

\begin{figure}[h]
\centering 
\includegraphics[width=1\linewidth]{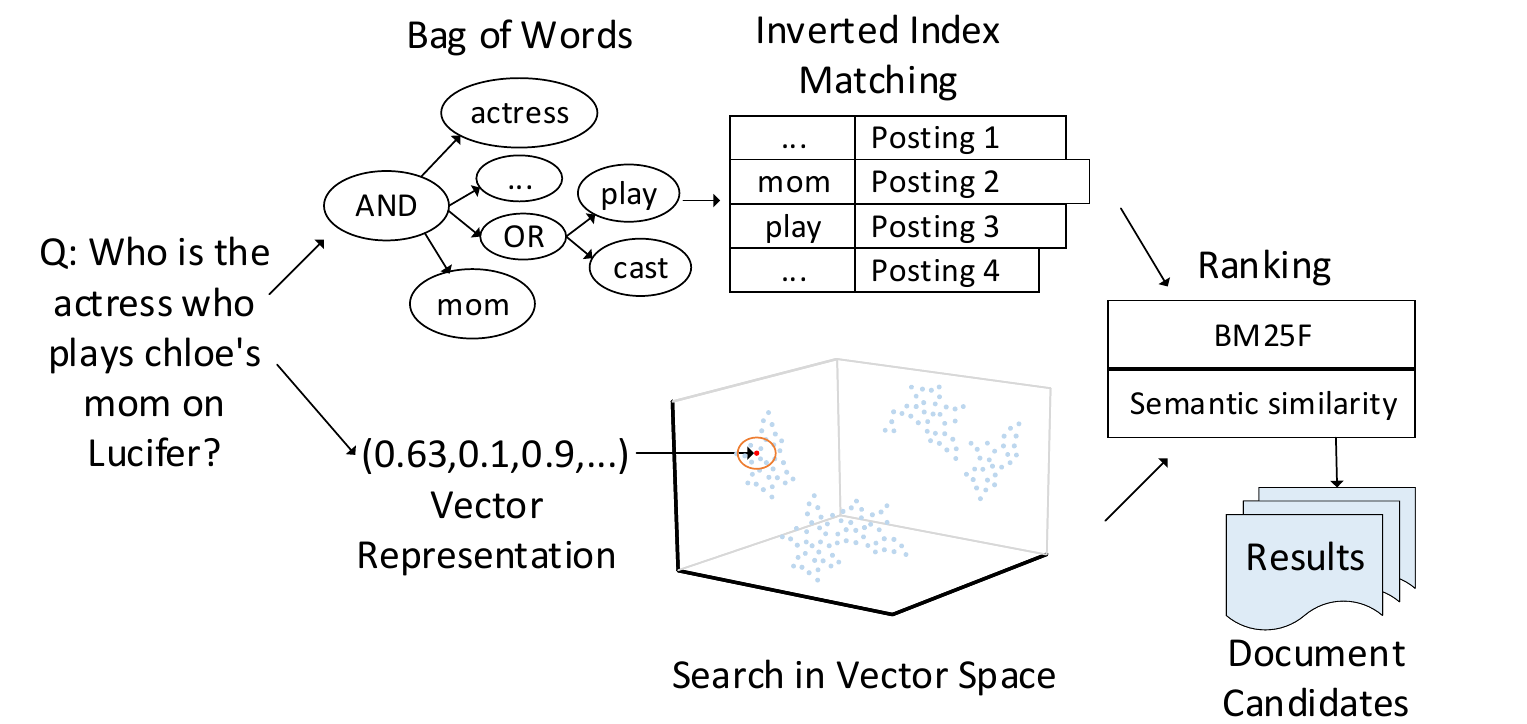}
\caption{Example of traditional web search vs. semantic based vector search.}
\label{fig:ann-ir-example}
\end{figure}

With the blooming of machine learning and deep learning, many information retrieval systems, such as web search, web question and answer, image search, and advertising engine, employ vector search to find the best matches to user queries based on their semantic meaning. 
Take web search as an example.  Fig.~\ref{fig:ann-ir-example} compares the traditional approach of web search to a semantic vector based retrieval.
Traditional search engines model similarity using bag-of-words and apply inverted indexes for keyword matching~\cite{google-search, bing-search, yahoo-search}, which is however difficult to capture the semantic similarity between a query and a document.
Thanks to the major advances in deep learning (DL) based feature vector extraction techniques~\cite{dssm, text-similarity-rnn-aaai16}, the semantic meaning of a document (or of a query) can be captured and encoded by a vector in high-dimensional vector space.
Finding semantically matching documents of a query is equivalent to a \emph{vector search} problem that retrieves the document vectors closest to the query vector. 
Major search engines such as Google, Bing, Baidu use vector search to improve web search quality~\cite{google-talk-to-books, bing-dl-image-search, baidu-semantic-search}.
Web search is just an example. 
Vector search is 
applicable to various data types such as images, code, tweets, video, and audio~\cite{ github-semantic-search}.

Searching for exact closest vectors is computationally expensive, especially when there are millions or even billions of vectors. In many cases, an approximate closest vector is almost as good as the exact one.
Because of that, approximate nearest neighbor search (ANN) algorithms are often used for solving the high-dimensional vector search problem~\cite{ann-survey}.

Real-world online services often pose stringent requirements on ANN search: 
high accuracy for effectiveness, low latency and small memory footprint for efficiency.
High accuracy is clearly important because the approximation has to return true nearest neighbors with high probability (\eg, $>0.95$, or $>0.99)$; otherwise, users will not be able to find what they are looking for and the service is useless. 
Low latency is crucial because online systems often come with stringent service level agreement (SLA) that requires responses to be returned within \emph{a few or tens of milliseconds}. Delayed responses could degrade user satisfaction and affect revenue~\cite{reduce-web-latency}. 
Small memory footprint is another important factor because memory is a scarce machine resource --- a memory efficient design is more scalable, e.g., reducing the number of machines needed to host billions of vectors, and empowering memory-constrained search on mobile devices and shared servers.

Popular existing approaches tackle ANN problems in high-dimensional space in two ways --- graph-based and quantization-based, both of which face challenges to offer desired accuracy, latency and memory all together.
The graph-based ANN approaches search nearest neighbors by exploring the proximity graph based on the closeness relation between nodes. They achieve high accuracy and  are fast, but they yield high memory overhead as they need to maintain both the original vectors and additional graph structure in memory~\cite{swg-ann,hnsw}.  Moreover, graph search requires many random accesses, which does not scale well on secondary storages such as SSDs.  
In a separate line of research, quantization-based ANN approaches, in particular product quantization (PQ) and its extensions, support low memory footprint index by compressing vectors into short code.  They, however, suffer from accuracy degradation~\cite{product-quantization, opq, lopq}, as the approximated distances due to compression cannot always differentiate the vectors according to their true distances to queries.  

In this paper, we tackle these challenges and propose an ANN solution, called \ANNeX, which collaboratively optimizes accuracy, latency and memory to offer both effectiveness and efficiency.
In particular, we build \ANNeX based on a \emph{multi-view} approach: a \emph{preview-step} with quantized vectors in-memory, which quickly generates a small set of candidates that have a high probability of containing the true top-$K$ NNs, 
and a 
\emph{full-view} step on SSDs to rerank those candidates with their full-length vectors. 
To improve efficiency, we design \ANNeX to reduce memory footprint through quantized representations and optimize latency through optimized routing and distance estimation; as for effectiveness, \ANNeX achieves high accuracy by reranking the close neighbors with full-length vectors to correctly identify the closest ones among them. 

We evaluate \ANNeX on two popular datasets \SIFT and \Deep.  We compare its effectiveness and efficiency with well-known ANN implementations including \IVFPQ and HNSW.  For efficiency, we not only measure classic metrics of latency and memory usage, we also propose and evaluate an ultimate efficiency/cost metric --- \emph{VQ}, i.e., VQ = number of \underline{V}ectors per machine $\times$ \underline{Q}ueries per second, inspired by DQ metric of web search engine~\cite{bitfunnel}. For a given ANN workload with a total number of vectors $Y$ and total QPS of $Q$, an ANN solution requires $Y \times Q / VQ$ number of machines.  
The higher the VQ, the less machines and cost!   


Our evaluation shows that, comparing with \IVFPQ, \ANNeX obtains significantly higher accuracy (from about 0.58--0.68 to 0.90--0.99), while improving VQ by 1.7--8.0 times.  
To meet similar accuracy target, we improve VQ by 12.2--14.9 times, which is equivalent to saving 12.2--14.9 times of infrastructure cost.
Comparing with HNSW, \ANNeX achieves 2.7--9.0 times VQ improvement with comparable accuracy.
As online services like web search host billions of vectors and serve thousands of requests per second through vector search, a highly cost-efficient solution like \ANNeX could save thousands of machines and millions of infrastructure cost per year.



The main contributions of the paper are summarized as below:
\begin{itemize}
\item Identify important effectiveness and efficiency metrics (\ie, VQ) of ANN search to offer cost-efficient high-quality online services. 
\item Develop a novel ANN solution --- \ANNeX --- that strikes for the highest VQ and thus lowest infrastructure cost while obtaining desired latency, memory, and accuracy.  
\item \ANNeX is the first ANN index that intelligently leverages SSD to reduce memory consumption while attaining compatible latency and accuracy.
\item Implement and evaluate \ANNeX: it collaboratively optimize latency, memory and accuracy, obtaining an order of magnitude improvements on VQ comparing with the state-of-the-art. 
\end{itemize}

\outline{Challenges.}

\outline{A set of techniques.}

\outline{Results}

\section{Background and Motivation}
\label{sec: background}

We first describe the vector search problem and then introduce the state-of-the-art ANN approaches. In particular, we describe NN proximity graph based approaches 
and quantization based approaches,
which are two representative lines.




\subsection{Vector Search and ANN}

The vector search problem is formally defined as:
\begin{defn} 
\emph{Vector Search Problem.} Let\ $Y = \{y_1,...,y_N\} \in \mathds{R}^D$ represent a set of vectors in a $D$-dimensional space and $q \in \mathds{R}^D$ the query. Given a value $K$, vector search finds the $K$ closest vectors in $Y$ to $q$, according to a pair-wise distance function 
$d \langle q, y \rangle$, as defined below:
\begin{equation}
TopK_q = \underset{y \in Y}{\operatorname{K-argmin}} \ d \langle q, y \rangle
\end{equation}
The result is a subset ${TopK_q} \subset Y$ such that (1) $|{TopK_q}| = K$ and 
(2) $\forall \ y \in Y - {TopK_q}, and\ y_q \in TopK_q: d(q, y_q) \leq d(q, y)$~\cite{k-nn-queries}.
\end{defn} 

In practice, the size $N$ of the set $Y$ is often large, so the computation cost of an exact solution is extremely high. To reduce the searching cost, approximate nearest neighbor (ANN) search is used, which returns the true nearest neighbors with high probability. 
Two lines of research have been conducted 
for high dimensional data: 
\emph{NN proximity graph based ANN} and 
\emph{quantization based ANN},
which are briefed below.



\subsection{NN Proximity Graphs}

The basic idea of NN proximity graph based methods is that a neighbor's neighbor is also likely to be a neighbor~\cite{nn-descent}. It therefore relies on exploring the graph based on the closeness relation between a node and its neighbors and neighbors' neighbors. Among those, the SWG (small world graph) approach builds an NN proximity graph  
with the small-world navigation property and obtains good accuracy and latency~\cite{swg-ann}.
Such a graph is featured by short graph diameter and high local clustering coefficient.
Yuri Malkov et al. introduced the most accomplished
version of this algorithm, namely HNSW (Hierarchical Navigable Small World Graph), which is a multi-layer variant of SWG.
Research shows that HNSW exhibits $O(\log N)$ search complexity 
($N$ represents the number of nodes in the graph), and performs well in high dimensionality~\cite{swg-ann, hnsw, small-world-kleinberg}. 

\notes{
HNSW selects a series of nested subsets of database vectors,
or "layers". The first layer contains only a single point, and
the base layer is the whole dataset. The sizes of the layers
follow a geometric progression, but they are otherwise sampled
randomly. For each of these layers HNSW constructs a
proximity graph. The search starts from the first layer.
A greedy search is performed on the layer until it reaches
the nearest neighbor of the query within this layer. That
vector is used as an entry point in the next layer as a seed
point to perform the search again. At the base layer, which
consists of all points, the procedure differs: a bread first
search starting at the seed produces the resulting neighbors. 
\minjia{No need to go into the details.}
}
HNSW and other NN proximity graph based approaches store the full length vectors and the full graph structure in memory, which incurs a memory overhead of $O(N \times D)$. Such a requirement causes a high cost of memory. 

\subsection{Quantization-based ANN Search}

Another popular line of
research for ANN search in high-dimensional space involves compressing high-dimensional vectors into short codes using product quantization and its extensions~\cite{product-quantization, opq, lopq}. 

\subsubsection{Product quantization}

Product quantization (PQ) takes high-dimensional vectors $y \in \mathds{R}^D$ as the input and splits them into
$M$ subvectors: $y = [y^1,...,y^M]$, where each $y^m \in \mathds{R}^{D'}, D' = D/M$.
Then these subvectors are quantized by $M$ distinct vector quantizers as:
$pq(y) = [vq^1(y^1),...,vq^M(y^M)]$. Each vector quantizer $vq^m$ has its own codebook, denoted by $C^m \subset \mathds{R}^{D'}$, which is a collection of $L$ representative \emph{codewords} $C^m = \{c^m_1,...,c^m_L\}$. The codebook $C^m$ can be built using Lloyd's algorithm~\cite{llyod-algorithm}.
The vector quantizer $vq^m$ maps each $y^m$ to its closest codeword in the codebook:
\begin{equation}
y^m \mapsto vq^m(y^m) = \underset{c^m_i \in C^m}{\operatorname{argmin}} \ d \langle y^m, c^m_i \rangle.
\end{equation}
The codebook of the product quantizer is the Cartesian product of the $M$ codebooks of those vector quantizers:
\begin{equation}
C_{pq} = C^1 \times ... \times C^M
\end{equation}
This method therefore could create $L^M$ codewords, but it only requires to store $L \times M \times D$ values for all its codebooks.

In practice, PQ maps the $m$-th subvecter of an input vector $y$ to an integer $(1,...,L)$, which is the index to the codebook $C^m$, 
and concatenates resultant $M$ integers to generate a \emph{PQ code} of $y$.
The memory cost of storing the PQ code of each vector is $B = M\ \times log_2{(L)}$ bits. Typically, $L$ is set to 256 
so that each subvector index is represented by one byte, and $M$ is set to divide $D$ evenly.

\subsubsection{Two-level quantization}
PQ and its extensions reduce the memory usage per vector.
However, the search is still exhaustive in the sense that the query is compared to all vectors. For large datasets, even reading highly quantized vectors is a severe performance bottleneck due to limited memory bandwidth. 
This leads to a two-level approach: at the first level, 
vectors are partitioned into $N_{cluster}$ clusters, represented by a set of cluster centroids $S_{centroid} = \{c_1,...,c_{N_{cluster}}\}$. Each $y$ then is mapped to its nearest cluster centroid through a mapping function $f_{cluster}$:
\begin{equation}
y \mapsto c_y = \underset{c_i \in C_{cluster}}{\operatorname{argmin}} \ d \langle y, c_i \rangle.
\end{equation}
The set of of vectors $V_i$ mapped to the same cluster is 
therefore defined as:
\begin{equation}
V_i = \{ y \in \mathds{R}^D : f_{cluster}(y) = c_i \}.
\end{equation}

At the second level, the approach uses a product quantizer $q_{pq}: \mathds{R}^D \mapsto C_{pq} \subset \mathds{R}^D$ to quantize the \emph{residual vector}, defined as the difference between $y$ and its  mapped cluster centroid $c_i$.
A database vector $y$ is therefore approximated as
\begin{equation}
y \approx f_{cluster}(y) + q_{pq}(y - f_{cluster}(y))
\end{equation}

Together we refer this two-level scheme as \IVFPQ\footnote{It is also sometimes called IVFADC, where IVF refers to Inverted File Index and ADC refers to the way of how the distance is calculated.}~\cite{product-quantization}.
While processing a query, \IVFPQ enables non-exhaustive searches, by finding either the closest or several of the closest \cells and scanning associated vectors only in those selected \cells at the second level~\cite{product-quantization}.

\section{Challenges}
\label{sec:challenges}



Both graph-based and quantization-based prior work faces challenges to offer desired accuracy, latency, and memory all together.


\paragraph{\textbf{Challenge I. Graph-based approaches are efficient, but they are memory consuming and do not scale well to SSDs.}} While graph-based approaches attain good latency and accuracy, their indices are memory consuming in order to store both full-length vectors and the additional graph structure ~\cite{hnsw}.  
As the number of vectors grows, its scalability is limited by the physical memory of the machine, and it becomes much harder to host all vectors on one machine.  A distributed approach would suffer from poor load balancing because the intrinsic nature of power-law distribution of queries~\cite{power-law-distributions-in-ir}, making scale-out inefficient.  On the other hand, SSDs have been widely used as secondary storage, and they are up to 8X cheaper and consume less than ten times power per bit than DRAM~\cite{energy-efficiency-survey, silt, rethink-dram-power-modes}. Many search engines such as Google and Bing already have SSDs to store the inverted index due to the factors
such as cost, scalability, and energy~\cite{search-engine-on-ssd, bing-ssd,search-engine-on-ssd, ssd-based-search-engine-architecture}. However, it is difficult to make HNSW work effectively on SSDs because the search of the graph structure generates many non-contiguous memory accesses, which are much slower on SSDs than memory and detrimental to the query latency.

\paragraph{\textbf{Challenge II. Quantization helps memory saving but results in poor recall.}}

The recall metric 
measures the fraction of the top-$K$ nearest neighbors retrieved by the ANN search which are exact nearest neighbors.
\footnote{Our recall definition is the same as the one used by HNSW. Quantization-based approaches often report $recall@R$, 
which is a different definition. It calculates the rate of queries for which the top-1 nearest neighbor is included in the top-$R$ returned results. The recall@1 is equivalent to the recall definition here when $K$ is 1. We are interested in \emph{recall} instead of \emph{recall@R} because in many scenarios it not only needs to know that the top-1 NN is included but also requires to identify which one is the top-1 NN.}

Quantization-based approaches, including \IVFPQ, suffer from low recall.
As a point of reference, with 64 times compression ratio, both \IVFPQ and its most advanced extension LOPQ achieve $< 0.4$ recall when $K=1$ ~\cite{product-quantization,lopq}. 

\begin{figure}[H]
\centering 
\includegraphics[width=0.7\linewidth]{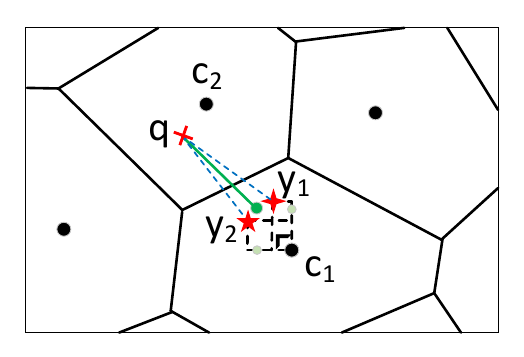}
\caption{2D \IVFPQ example of the lossy procedure of quantization-based approaches. $q$ denotes the query, $y_1$ and $y_2$ (drawn as a red cross) denotes vectors. $c_1$ and $c_2$ represent cluster centroids. 
}
\label{fig:pq-example}
\end{figure}

\begin{figure}[H]
\centering 
\includegraphics[width=1\linewidth]{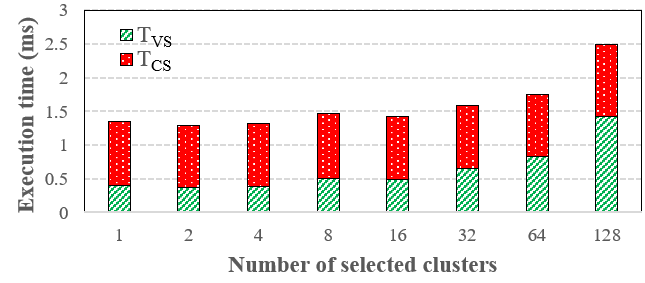}
\caption{\SIFT dataset: Evaluation of \cell selection time $T_{CS}$ and vector scanning time $T_{VS}$ with two-level quantization-based approach. The x-axis represents the number of selected \cells out of a total of 16K clusters.}
\label{fig:sift1m-latency-large-k-varying-cluster} 
\end{figure}

To see the reason behind why \IVFPQ and in general quantization-based approaches incur low recall, especially with large compression ratio, let us see a simple example as illustrated in Fig.~\ref{fig:pq-example}. 
\noindent
Assume we have a 2D vector space with two vectors $y_1$ and $y_2$ assigned to the same first-level \cell represented by $c_1$.
Because the quantization process is lossy, $y_1$ and $y_2$ might be quantized to have the same PQ code (represented by the green dot). Their distance to a query $q$ is the same, 
and it is impossible to differentiate which one is closer to the query using the quantized vectors.
As a result, quantization generate non-negligible recall loss.
\paragraph{\textbf{Challenge III. Quantization-based approaches face a dilemma on optimizing latency.}}
As one-level quantization suffers from exhaustive scan, we focus on two-level quantization, whose latency is decomposed into two parts: the \emph{cluster selection time} (\textbf{$T_{CS}$}) and the \emph{vector scanning time} (\textbf{$T_{VS}$}) in selected \cells. Both latency components depend on the number of the first-level clusters \Ncluster.
During cluster selection, an exhaustive search is used to find out a few \cells closest to the query to scan, and 
$T_{CS}$ in this case is linear in \Ncluster. When \Ncluster is not too large, 
finding which \cell to scan is computationally inexpensive. Existing approaches rarely choose a large $N_{cluster}$, because as $N_{cluster}$ increases, $T_{CS}$ itself would become too long, prolonging  query latency.\footnote{Another reason is perhaps that standard clustering algorithms are prohibitively slow when the number of clusters is large.} 
Fig.~\ref{fig:sift1m-latency-large-k-varying-cluster} shows that for 1M vectors with \Ncluster $= 16K$, $T_{CS}$ already takes a significant portion of execution time than $T_{VS}$.\footnote{Evaluation is conducted on \SIFT dataset.
}


\begin{figure}[H]
\centering 
\subfloat[Evaluation of recall of two-level quantization. All three configurations can reach almost the same recall by scanning the same number of \cells.]{\label{fig:sift1m-recall-varying-cluster}\includegraphics[width=1\linewidth]{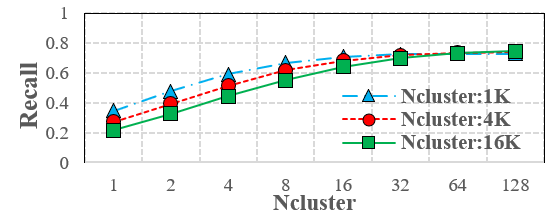}} \\
\subfloat[Plot of the percentage of selected \cells to scan. Given the same number of \cells to select, larger \Ncluster requires a smaller percentage of \cells to be scanned.]{\label{fig:sift1m-percentage-varying-cluster}\includegraphics[width=1\linewidth]{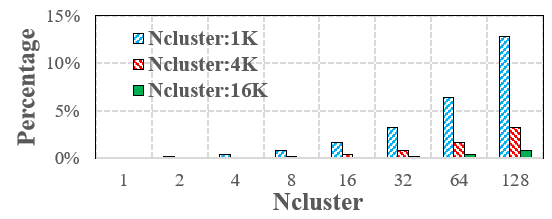}} \\
\caption{\SIFT dataset: Evaluation of recall and the percentage of scanned \cells with two-level quantization for three different number of \cells $N_{cluster}$. The x-axis is the number of selected \cells to scan.}
\label{fig:sift1m-perf-varying-cluster} 
\end{figure}

On the other end, 
larger \Ncluster leads to a smaller percentage of vectors to be scanned at the second level to reach the same recall. Fig.~\ref{fig:sift1m-recall-varying-cluster} shows that given three different values of \Ncluster 1K, 4K, and 16K, scanning the same number of \cells (\eg, 64) all lead to close or very similar recall. 
This observation is kind of intuitive as top $K$ NNs would belong to at most $K$ clusters regardless of \Ncluster value.
As larger \Ncluster has less (expected) number of vectors per cluster, this observation indicates that larger $N_{cluster}$ requires less number of vectors to be scanned at the second level. 
Figure~\ref{fig:sift1m-percentage-varying-cluster} confirms that by scanning 64 \cells, only 0.4\% cells need to be scanned when $N_{cluster}$ is $16K$ whereas it is 6.4\% when $N_{cluster} = 1K$, which means the search at the second level can take much longer time with smaller $N_{cluster}$.

The $N_{cluster}~$ dilemma makes it challenging to optimize both $T_{CS}$ and $T_{VS}$.  The problem is beyond selecting a good value for $N_{cluster}$ ; it requires more fundamental change on ANN design for latency optimization.

\section{Design Overview}
\label{sec: design}


In this section, we propose an ANN search solution to address the aforementioned challenges, offering low latency, high memory efficiency, and high recall all together.
The software architecture overview is presented in Fig.~\ref{fig:overview}.

First, a key \textit{empirical observation} enlightened us to solve the poor recall challenge of quantization such that we can achieve high memory efficiency and high recall together --- 
\emph{Although the \textbf{approximated} top-$K$ NNs might not always match the \textbf{exact} top-$K$ NNs due to the precision loss from quantization,
the approximated top-$K$ NNs are more likely to fall within a list of top-$R$ candidates, where $R$ is larger but not too much larger than $K$.} 

\begin{figure}[H]
\centering 
\subfloat[Probabilties of top-1 NN falls within top-R candidate results. Although the approximated top-1 might not always match the exact top-1 NN, the top-1 NN has a high chance to be within the top-10 results.]{\label{fig:sift1m-demo-R1}\includegraphics[width=0.9\linewidth]{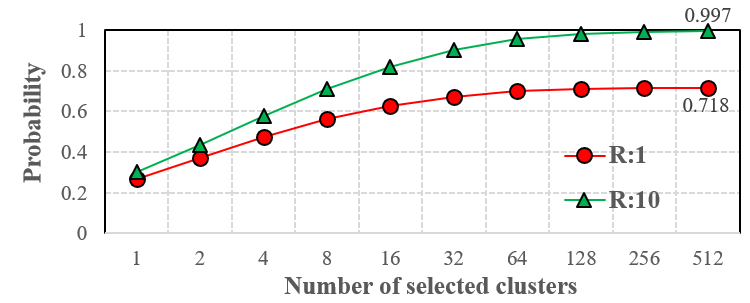}} \vspace{-1.2em}
\subfloat[Probabilties of top-10 NNs fall within top-R candidate results. The approximated top-10 results are more likely to be within the top-50 candidates.]{\label{fig:sift1m-demo-R10}\includegraphics[width=0.9\linewidth]{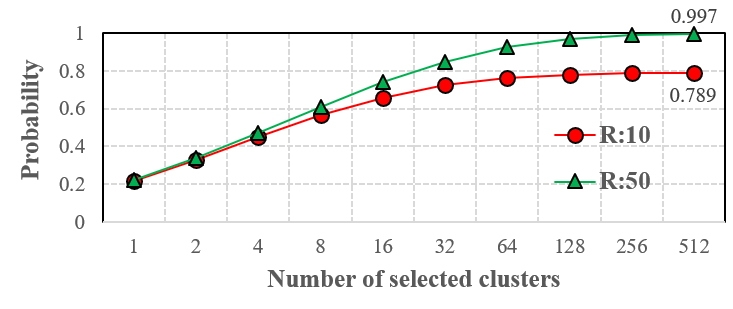}} \\
\caption{A key observation from quantization recall analysis. 
}
\label{fig:top-k-in-top-r} 
\end{figure}

\begin{figure}[H]
\centering 
\includegraphics[width=1\linewidth]{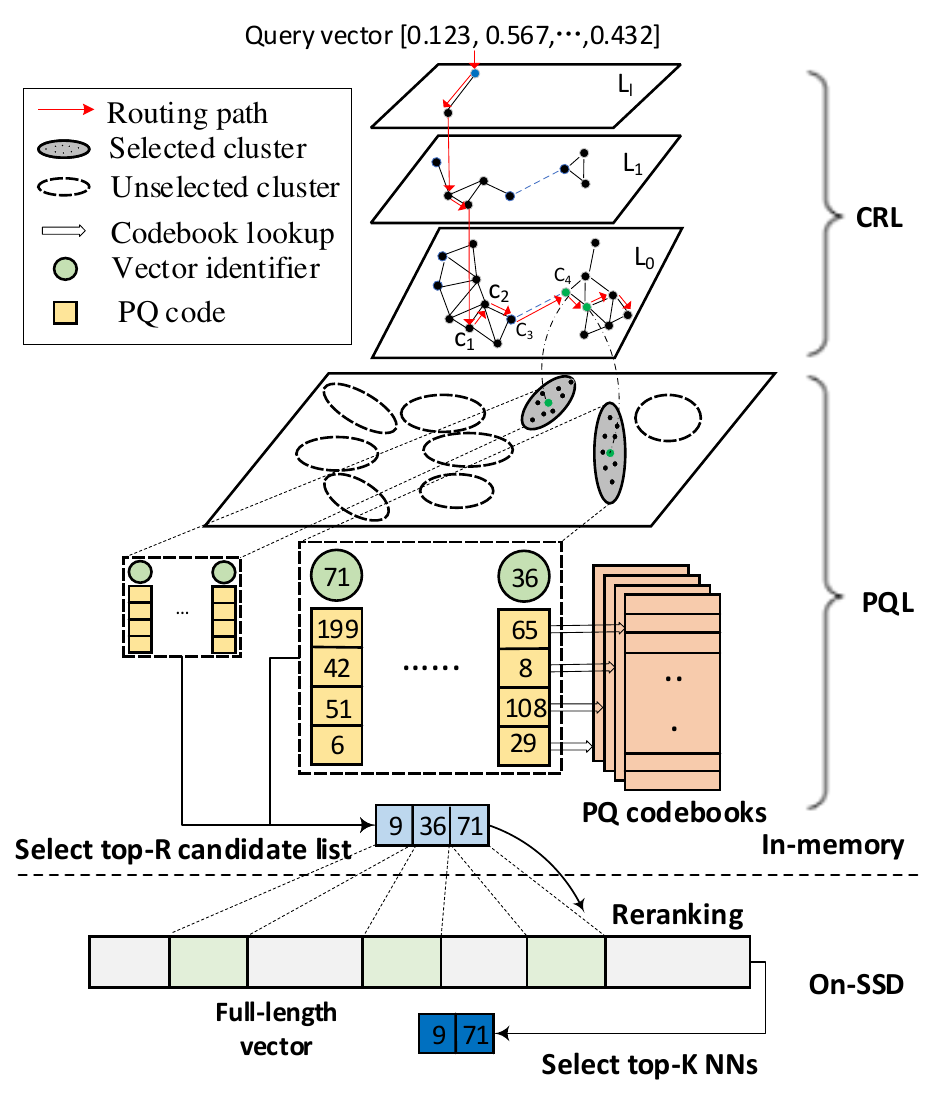}
\caption{The \ANNeX architecture.}
\label{fig:overview}
\end{figure}

\paragraph{Observation on quantization recall.} We made the observation through recall analysis of \IVFPQ, and we found it general towards various datasets (more results in Section~\ref{subsec:eval-dq}). 
Fig.~\ref{fig:sift1m-demo-R1} shows an example on \SIFT dataset, where we apply \IVFPQ by partitioning the vector space into $16K$ clusters and quantizes vectors with a compression ratio of 16X. By scanning 512 clusters (~3\% of the total clusters), the likelihood of finding the top-1 NN in top-10 candidates is 99.7\%, whereas the probability of an exact match is only 71.8\%. 
Similarly, Fig.~\ref{fig:sift1m-demo-R10} shows that the probability of finding the top-10 NNs in top-50 candidates is close to 1, whereas the recall (\eg, when R=10) is only 78.9\%. This is presumably because quantization makes it impossible to differentiate true NNs from their neighbor vectors if they are quantized to have the same PQ code, as shown in Fig.~\ref{fig:pq-example} in Section~\ref{sec:challenges}. Therefore, the true NNs are included in top-$R$ but cannot be identified due to the precision loss. 
Our analysis indicates that such relation holds for $K > 1$ as well. 

\later{
Furthermore, non-volatile solid-state drive (SSD) and other emerging storage technologies  are poised to close the enormous performance gap between hard
disk drives (HDDs) and main memory. SSDs can achieve a 2–3 orders of
magnitude increase in I/O operations per second (IOPS) and decrease in latencies over HDDs. Commercial available SSDs~\footnote{https://www.samsung.com/semiconductor/minisite/ssd/product/consumer/960pro/} can produce up to 440,000 read read accesses with large storage capacity and low power consumption~\cite{rethink-flash-in-data-center, flash-cost}. Although SSDs are still more expensive than HDDs, they have been deployed widely in both desktop~\cite{ssd-in-desktop} and enterprise~\cite{facebook-ssd,oracle-ssd} environments due to their much lower access latency and declining costs. 
However, today's ANN algorithms have long been focused on building in-memory index, leading to the question: can we leverage SSDs to make ANN index more scalable?
}

\paragraph{Overview.}
Based on the observation,
we propose an ANN design, called \ANNeX, that employs a "multi-view" approach 
as an \textbf{accuracy enhancement procedure}, contrasting it with a single-view approach (either quantized representation, as in \IVFPQ, or full-length vectors, as in HNSW). The multi-view approach contains two steps: 
\begin{itemize} 
	\item A \textbf{\emph{preview} step in memory} that employs a preview index as a filter to generate a small candidate set of top-$R$ NNs based on quantized vectors;
    \item A \textbf{\emph{full-view} step on SSDs} that reranks the selected NNs from the preview step based on their full-length vectors and selects top-$K$ NNs from the top-$R$ candidates.
\end{itemize}



Intuitively, such a multi-view design achieves memory savings, through in-memory preview on quantized data, and high recall, through SSD-based reranking on full-length vectors, but what about latency? The full-view step only needs to rerank a small set of candidates, so this part of latency is less of a  big concern, but what about the preview step? 
To address the latency challenge of quantization, the preview step of \ANNeX is powered by 
a \emph{cluster routing layer (CRL)} together with a \emph{product quantization layer (PQL)}. The \cell routing layer 
leverages HNSW to quickly and accurately dispatch a query to a few nearest \cells, instead of scanning centroids of all clusters.  HNSW reduces the cluster selection cost from $O(N_{cluster})$ to $O(\log(N_{cluster}))$, and thus, \ANNeX can choose a relatively large value of $N_{cluster}$ with affordable $T_{CS}$ and small $T_{VS}$, addressing Challenge III and optimizing latency.    
The product quantization layer compresses vectors through product quantization with optimized distance computation that improves overall system efficiency.

Next, we describe \ANNeX index construction in Section~\ref{sec:index-construction} and query processing in Section~\ref{sec:ann-search}. 



\later{
\begin{table}[]
\centering
  	\newcommand{\colspc}{\hspace*{0.43em}}
    \newcommand{\smallercolspc}{\hspace*{0.33em}}
    \small
    \renewcommand\sfsmaller{}
\begin{tabular}{@{\colspc}c@{\colspc}c@{\colspc}c@{\colspc}}
Symbol & Meaning                              & Example \\ \hline
       & \underline{Workload characteristics} &         \\
N      & number of vectors           & 1M      \\
D      & number of dimensions of input vector                 & 128     \\
f      & number of bytes of data type         & 4      \\ 
       & \underline{System parameters}         &         \\
\Ncluster     & number of first-level \cells        & $2^{15}$  \\
\Npq     & number of sub-codeword per sub-dimension & $2^8$    \\
M      & number of sub-dimension              & 32      \\
$N_{scan}$ & number of \cells to scan at the second-level  & 256     \\
R	   & number of candidate NNs at the preview step  & 50  \\
K	   & number of nearest neighbors to retrieve  & 10     \\ \hline
\end{tabular}
\caption{Notation.}
\label{tbl:notation}
\end{table}
}

\section{Index Construction}
\label{sec:index-construction}





Given a set of vectors, \ANNeX builds the in-memory \emph{\textbf{preview index}}  using 
Algorithm~\ref{algo:index-build-step}, with 3 major steps: 
\begin{itemize}
\item \textbf{Clustering}. Cluster on the set of data vectors $Y$ to divide the space into \Ncluster \cells. \ANNeX keeps the set of centroids of those clusters at \Ccluster
(line~\ref{lst:preview-step:clustering}).
\item \textbf{Routing layer construction}. Use HNSW to build a routing structure on top of \Ccluster to quickly identify a few closest \cells to scan (line~\ref{lst:preview-step:routing-layer}). Perform \emph{connectivity augmentation} to ensure the reachability of \cells (line~\ref{lst:preview-step:connectivity}). 
\item \textbf{PQ layer generation}. The PQ layer leverages product quantization to generate PQ codebooks and PQ encoded vectors~\cite{product-quantization}. To generate PQ codebooks, \ANNeX first calculates the residual distance of each vector to the \cell it belongs to (line~\ref{lst:preview-step:compute-residual-begin}--line~\ref{lst:preview-step:compute-residual-end}). It then splits the vector space into $M$ sub-dimensional spaces and constructs a separate codebook for each sub-dimension (line~\ref{lst:preview-step:train-residual-begin}--line~\ref{lst:preview-step:train-residual-end}). Thus, \ANNeX has $D/M$ dimensional codebooks for $M$ sub-dimension, each with $L$ subspace codewords. Generating the codebooks based on residual vectors is a common technique to reduce the quantization error~\cite{multi-stage-vector-quantization, product-quantization}.
\ANNeX then generates PQ encoded vectors by assigning the PQ code for each vector (\ie, a concatenation of $M$ indices) and adds each quantized vector to the \cell it belongs to (line~\ref{lst:preview-step:assignment-begin}--line~\ref{lst:preview-step:assignment-end}). Furthermore, 
\ANNeX precomputes terms used for the PQ code distance computation and store the precomputed results into a cache (line~\ref{lst:preview-step:precomputation-begin}--line~\ref{lst:preview-step:precomputation-end}). Such a precomputation helps reduce memory accesses during the search phase to obtain lower latency. 
\end{itemize}

\noindent
\emph{\textbf{Full-view index}}. \ANNeX stores and uses full-length vectors for reranking. It 
keeps vectors as binary vectors byte-aligned as a file on SSDs (line~\ref{lst:fullview-step:strore}), which allows each vector to be selected through random access (\eg, based on vector id/offset). Using HDDs is detrimental to query latency because the slow, mechanical sector seek time for random accesses.

\begin{algorithm}[h!]
\caption{\hfill \textbf{\ANNeX index construction algorithm}}
\label{algo:index-build-step}
  \begin{algorithmic}[1]
    \State \textbf{Input:} Vector set $Y$, vector dimension $D$.
    	\label{lst:preview-step:input}
    \State \textbf{Output:} \ANNeX \emph{index}.
    	\label{lst:preview-step:output}
    \State \textbf{Parameter:} Number of sub-dimensional spaces $M$, number of sub-codewords $L$ in each sub-codebook, size of clusters \Ncluster.
    	\label{lst:preview-step:hyperparameter}        
    \State $index.S_{centroid} \leftarrow clustering(Y, D, N_{cluster})$
		\Comment{Partition the vector space using K-Means algorithm.}
        \label{lst:preview-step:clustering} 
    \State $index.routing\_layer \leftarrow CreateHnsw(S_{centroid}, M)$
    	\label{lst:preview-step:routing-layer} 
    \State $connectivity\_augmentation(routing\_layer)$
    	\label{lst:preview-step:connectivity} 
    \ForAll {$i$ \textbf{in} 0..($N-1$)}
    	\label{lst:preview-step:compute-residual-begin}
    	\State $cell\_id, codeword \leftarrow assign(Y[i], S_{centroid})$ 
    	\State $R[i] \leftarrow compute\_residual(Y[i], codeword)$ 
        \State $CID[i] \leftarrow cell\_id$
        \label{lst:preview-step:compute-residual-end} 
    \EndFor    
    \ForAll {$i$ \textbf{in} 0..($M-1$)}
    	\label{lst:preview-step:train-residual-begin} 
        \State $codebook \leftarrow train\_residual(R[:i \times D/M, (i + 1) \times D/M], L))$ 
    	\State $pq\_codebook.set(i,  codebook)$ 
        \label{lst:preview-step:train-residual-end} 
    \EndFor
    \State $index.pq\_layer.add(pq\_codebook)$ 
    \ForAll {$i$ \textbf{in} 0..($N-1$)}
		\label{lst:preview-step:assignment-begin}
		\State $cell\_id \leftarrow CID[i]$
        \State $Y_{pq\_code}[i] \leftarrow product\_quantizer(R[i])$
    	\State $pq\_layer.lists[cell\_id].add(Y_{pq\_code}[i])$
        \label{lst:preview-step:assignment-end} 
    \EndFor  
    \ForAll {$i$ \textbf{in} 0..($N-1$)}
    	\label{lst:preview-step:precomputation-begin} 
         \State $index.cache[i] \leftarrow precompute\_term(R[i], CID[i])$
        	\label{lst:preview-step:precomputation-end} 
    \EndFor
    \State $store\_fullview\_vectors()$ 
         \label{lst:fullview-step:strore} 
  \end{algorithmic}
\end{algorithm}






\subsection{\textbf{Routing Layer Construction}}
\label{subsec:hnsw}
Performing exact search to find the closest clusters incurs complexity of $O(N_{cluster} \times D)$. 
We explore HNSW as a routing scheme to perform approximate search, achieving $O(\log (N_{cluster}) \times D)$ complexity while attaining close to unity recall with fast lookup~\cite{swg-ann, hnsw, small-world-kleinberg}.

\paragraph{HNSW index for routing}
We describe the main ideas and refer readers to ~\cite{hnsw} for more details.
The routing layer is built incrementally by 
iteratively inserting each centroid $c_i$ of $S_{centroid}$ 
as a node. 
Each node generates $OutD$ (\ie, the neighbor degree) out-going edges. Among these, $OutD - 1$ are \emph{short--range} edges, which connect $c_i$ to $OutD - 1$ closest centroids according to their pair-wise Euclidean distance to $c_i$ (\eg, the edge between $c_1$ and $c_2$ in Fig.~\ref{fig:overview}). The rest is a \emph{long--range} edge that connects $c_i$ to a randomly picked node, which does not necessarily connect two closest nodes but may connect other locally connected node clusters (\eg, the edge between $c_3$ and $c_{4}$ in Fig.~\ref{fig:overview}). It is theoretically justified that constructing a proximity graph by inserting these two types of edges offers the graph small-world properties~\cite{small-world-dynamics,swg-ann, hnsw}.

The constructed small world graph using all $c_i$ becomes the ground layer $L_0$ of CRL.
\ANNeX then creates a hierarchical small world graph by creating a chain of subsets $V = L_0\supseteq L_1\supseteq\ldots\supseteq L_l$  of nodes as "layers", where
each node in $L_{i}$ is randomly selected to be in $L_{i+1}$ with a fixed probability $1/OutD$.
On each layer, the edges are defined so that the overall structure becomes a small world graph, and the number of layers is bounded by $O\left(\log (N_{cluster})/\log (OutD)\right)$~\cite{swg-ann}. 

\begin{figure}[H]
\centering
\includegraphics[width=0.7\linewidth]{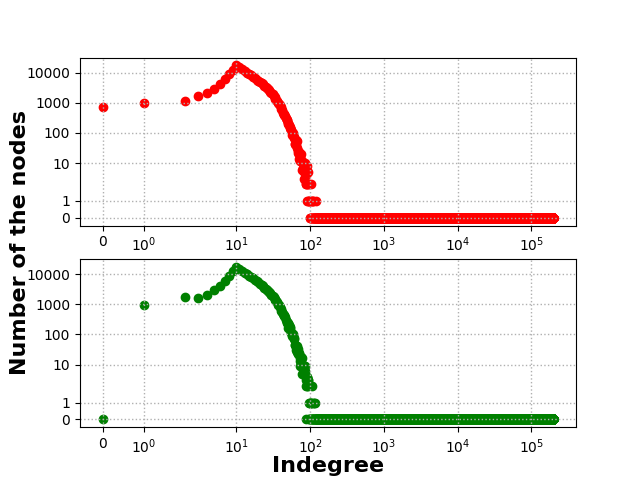}
\caption{Indegree histogram of the HNSW routing layer.
}
\label{fig:graph_connection_file_M10_efC200}
\end{figure}

\paragraph{Connectivity augmentation.}
HNSW does not guarantee the reachability of all nodes. There can be a large amount of "isolated nodes" at the ground layer, which are nodes with
zero in-degree.
When used as a routing scheme, it is crucial to ensure reachability 
as an entire \cell
will be missed if HNSW cannot reach 
its centroid.
Fig.~\ref{fig:graph_connection_file_M10_efC200} shows the frequency distributions of in-degree for all nodes in the routing layer~\footnote{Results are obtained on 200K cell centroids of Deep10M dataset.}. Without optimization (Fig.~\ref{fig:graph_connection_file_M10_efC200} (top)), there are around 1,000 nodes whose in-degree are zero. These nodes and their corresponding \cells cannot be reached during the search process. 

We develop a new technique called \emph{connectivity augmentation} to resolve the issue. We apply Kosaraju's algorithm~\cite{strong-connectivity-survey} against the constructed routing layer.
This step (line~\ref{lst:preview-step:connectivity}) adjusts the routing layer by adding minimal number of edges to make the graph strongly connected without destroying its small world properties. Fig.~\ref{fig:graph_connection_file_M10_efC200} (bottom) shows that after connectivity augmentation, there are no zero in-degree nodes.

\subsection{\textbf{Towards Larger \Ncluster}}
As shown in Fig.~\ref{fig:sift1m-perf-varying-cluster}, the number of vectors to be scanned for a target accuracy is strongly determined by \Ncluster. 
Choosing a large \Ncluster 
reduces that, so does the \cell scanning time.

A major challenge of having a large \Ncluster is the \cell selection time $T_{CS}$, which we address through the HNSW based routing. 
Another challenge is that the standard K-Means algorithm is prohibitively slow. 
To boost the speed of clustering with large \Ncluster, we employ \emph{Yingyang $K$-Means} and GPU to perform the clustering, which speedup normal $K$-Means by avoiding redundant distance computation if cluster centroids do not change drastically in between iterations~\cite{yingyang-kmeans}~\footnote{https://github.com/src-d/kmcuda}. One nice property of this approach is that it serves as a drop-in replacement and yields exactly the same results that would be achieved by ordinary $K$-Means.

Memory consumption is another concern, yet even a large \Ncluster is still relatively small compared to the quantized representation of the entire set of vectors. The memory cost of \ANNeX index is given by: 
\begin{equation} \label{eq5}
\begin{split}
MO & = N \times( M \times \frac{log(L)}{8\textnormal{-}bit} + f)  + L \times D \times f \\
   & + N_{cluster} \times (D + OutD) \times f \ ,
\end{split}
\end{equation}
which is the sum of the size of cluster centroids ($N_{cluster} \times D \times f$), metadata for the routing layer ($N_{cluster} \times 2 \times OutD \times f$), the PQ codebooks ($L \times D \times f$ bytes), and the PQ code ($M \times \frac{log(L)}{8\textnormal{-}bit}$) plus $f$-byte per vector for caching the precomputation result, where $f$ denotes the number of bytes of vector data type.

\later{
The compressed vector layer compresses vectors into short-code. Existing IVFADC computes distance by breaking down the ADC into multiple terms and relies on multiple look-up operations (LUTs) to compute distances. This design focuses on getting low memory footprint, but it amplifies the memory bandwidth consumption problem creating a key performance bottleneck. Since the memory bandwidth (\ie, the rate at which data is moved from slow cache to fast cache) is a very limited resource, having too many LUTs can limit the performance and scalability of ANN search. }

\later{
\subsubsection{\textbf{Optimizing distance computation}}
The ADC is a memory intensive operation, and its performance is limited by memory bandwidth.
Memory bandwidth is a constraining resource on modern CPUs. For instance, the \emph{peak} computational performance of a Xeon E5-2650 machine is \peakperf while the observable $DataBandwidth$ between $L3$ and $L2$ cache on it is $62.5$ GigaFloats/s ($250$GB/s)~\footnote{Measured using the stream benchmark~\cite{stream}}, which is much smaller. If the total number of LUTs is high, the total execution time is dominated by the data movement, resulting in poor performance and preventing ANN index from fully exploiting the power of the hardware. Here, we present a partial ADC caching method that significantly reduces memory accesses during the search time and mitigates memory bandwidth consumption. 
}

\section{Query Processing}
\label{sec:ann-search}

We describe how \ANNeX searches top-$K$ NNs.
Algorithm~\ref{algo:preview-step-search} shows the online search process. Preview search happens in memory, starting from the routing layer, which returns the \emph{ids} ($I^{Nscan}$) and distance ($D^{Nscan}$) of $Nscan$ selected \cells (line~\ref{lst:preview-search:routing}). For those selected \cells, \ANNeX scans vectors in each of them at the PQ layer and keeps track of the top-$R$ closest candidate NNs (line~\ref{lst:preview-search:generate-candidate-set}). 
The top-$R$ candidates are reranked during full view (line~\ref{lst:search:rerank-with-fullview}). 

\begin{algorithm}[h!]
\caption{\hfill \textbf{\ANNeX online search algorithm}}
\label{algo:preview-step-search}
  \begin{algorithmic}[1]
    \State \textbf{Input:} Query vector $q$, number of nearest neighbors $K$ to retrieve.
    	\label{lst:preview-step:input}
    \State \textbf{Output:} Top-$K$ nearest neighbors.
    	\label{lst:preview-step:output}
    \State \textbf{Parameter:}  number of \cells to scan $Nscan$, size of HNSW search queue $efSearch$, size of candidate list $R$.
    	\label{lst:preview-step:hyperparameter}
    \State $I^{Nscan}, D^{Nscan} \leftarrow index.routing\_layer.search(q, Nscan)$
		\Comment{ Return top-$Nscan$ \cells with minimal distance to $q$.}
        \label{lst:preview-search:routing} 
    \State $TopR \leftarrow scan\_pq\_vectors(q, I^{Nscan}, D^{Nscan})$ 
        \Comment{Preview step.}
    	\label{lst:preview-search:generate-candidate-set}
    \State $TopK \leftarrow rerank\_with\_fullview\_vectors(q, TopR)$         \Comment{Full-view step.}
    	\label{lst:search:rerank-with-fullview}
  \end{algorithmic}
\end{algorithm}

\subsection{Preview Step}
\label{subsub: preview-step-search}

\textbf{Clusters selection.} \ANNeX selects \cells using the search method from HNSW~\cite{hnsw}, which is briefly described below. The search starts from the top of the routing layer
and uses greedy search 
to find the node with the closest distance to the query $q$ as an entry point to descend to the lower layers. The upper layers route $q$ to an entry point in the ground layer that is 
in a region close to the nearest neighbors to $q$. Once reaching the ground layer, \ANNeX employs prioritized breath-first search: It exams its neighbors and stores all the visited nodes in a priority queue based on their distances to the $q$. The length of the queue is bounded by $efSearch$, a system parameter that controls the trade-off between search time and accuracy.
When the search reaches a termination condition (e.g., the number of distance calculation), \ANNeX returns $Nscan$ closest \cells. 


\paragraph{\textbf{PQ layer distance computation.}}
\label{subsubsec: caching-adc}
\ANNeX calculates the distance from query $q$ to a data point $y$ in cluster $V_i$ using its PQ code and asymmetric distance computation (ADC)~\cite{product-quantization}: 
\begin{eqnarray}
d\langle q, y \rangle \!\!\!&=&\!\!\!\! d\langle q - c, r \rangle \sim d_{ADC}\langle q-c, pq(r) \rangle \\
\!\!\!&=&\!\!\!\! \sum_{m = 1}^{M}{d\langle (q-c)^m, vq^m(r^m) \rangle} 
\label{eqn:adc-dist}
\end{eqnarray}
where $c$ is the cluster center and $r$ is the residual distance between $y$ and $c$, represented as $[r^1:,,,:r^M]$. 
\later{
A basic approach to scan selected \cells is therefore to calculate the ADC distance on-the-fly between the query and every vector and keeps track of $R$ closest candidates, as shown in the $basic\_scan\_vectors()$ method in Lst.~\ref{lst:naive-dist-computation}.
}
\later{
\begin{lstlisting}[caption=Basic way of scanning preview vectors, label=lst:naive-dist-computation, mathescape=true] 
basic_scan_preview_vectors(q, $I^{Nscan}$, $D^{Nscan}$)
    TopR = min_priority_queue()
    for n in 0..($Nscan-1$):
        cluster = index.compressed_layer.list$[I^{Nscan}[n]]$  
    	    for r in cluster:            
                dist = $d_{ADC} \langle (q-D^{Nscan}[n]), r \rangle$             
                TopR.push(r, dist)
    return TopR
\end{lstlisting}
}
\later{
The basic approach requires $M$ lookups to load the product quantized $r$ value and $O(D)$ arithmetic operations to calculate the distance for each vector. 
}
It can be expanded into four terms for Euclidean norm calculation~\cite{improve-bilayer-ann}: 
\begin{eqnarray} 
\underset{Term-A}{\underline{\|x - c\| ^2}} + \underset{Term-B}{\underline{\sum_{m = 1}^{M}{\|c_{y^m}^m\|}^2}} + \underset{Term-C}{\underline{2 \sum_{m = 1}^{M}{\langle c^m, c_{y^m}^m \rangle}}} - \underset{Term-D}{\underline{2 \sum_{m = 1}^{M}{\langle x^m, c_{y^m}^m \rangle}}} 
\label{eqn:adc-expanded-terms}
\end{eqnarray}
where $c_{y^m}^m = vq^m(r^m)$, denoting the closest sub-codeword assigned to sub-vector $y^m$ in the $m$-th sub-dimension.



Existing approach calculates these terms on-the-fly for each query by calculating them and 
reusing the results with
look-up tables (LUTs). Without optimization, it requires $2 \times M$ lookup-add operations to estimate the distance per data point.

By looking close into these terms,  
we notice that each term can be 
query dependent (query-dep.), mapped centroid dependent
(centroid-dep.), and/or PQ code dependent (PQ-code-dep.). Table~\ref{tbl:adc-term} summaries these dependencies.

\begin{table}[H]
\small
\begin{tabular}{|c|c|c|c|}
\hline
       & \textbf{Query-dep.} & \textbf{Centroid-dep.}& \textbf{PQ-code-dep.} \\ \hline
Term-A & \cmark               & \cmark              & \xmark                  \\ \hline
Term-B & \xmark               & \xmark              & \cmark                  \\ \hline
Term-C & \xmark               & \cmark              & \cmark                 \\ \hline
Term-D & \cmark               & \xmark              & \cmark                  \\ \hline
\end{tabular}
\caption{Dependencies of
PQ distance computation.}
\label{tbl:adc-term}
\end{table}

\later{
Among these four terms, 
Term-A only needs to be computed \emph{once for each cluster}. 
Term-D depends on both the query and the PQ codebooks. 
Since for each m-th sub-dimension, there are only $L$ possible codewords, the corresponding $\langle x_m, c_{y^m}^m \rangle$ distance only needs to be computed \emph{once for each query}.
}

Since both Term-B and Term-C are query independent,
\ANNeX precomputes Term-B and Term-C 
and caches their sum 
for each data point as a post-training step. During query processing, it takes a single lookup to get the sum of Term-B and Term-C. It then requires only $M + 1$ lookup-add operations to estimate the distance, with the trade-off of adding $f$-byte memory (\eg, 4-byte if vector type is float) per data point.
The $scan\_pq\_vectors()$ method in Lst.~\ref{lst:dist-computation} shows how \ANNeX scans a \cell with the optimized distance computation. 

\begin{lstlisting}[caption=Method to scan PQ vectors, label=lst:dist-computation, mathescape=true] 
scan_pq_vectors(q, $I^{Nscan}$, $D^{Nscan}$)
    TopR = min_priority_queue()
    // init LUTs for computing all possible Term-D
    for m in M:
        for l in L:
    	    termD_LUT[m][l] = -2 x $\langle x^m, c_l^m \rangle$  
    for n in 0..($Nscan-1$):
        cluster = index.pq_layer.list$[I^{Nscan}[n]]$
            // Compute Term-A
            t1 = $\|q - D^{Nscan}[n]\| ^2$
    	    for v in cluster: 
                // M lookup-add Term-D
                for m in M:
                    dist += termD_LUT[m]
                // Lookup-add precomputed Term-B and Term-C in the index construction phase
                dist += index.cache[v]
                TopR.push(v, dist)
    return TopR
\end{lstlisting}

\subsection{Full-view Step}
\label{subsec: post-validation-search}

\outline{What is post-validation or post-verification?}

\outline{How this post-validation phase is traditionally done? Accessing disk is expensive. Often report recall@R, which means the top-most NN exists in top-R candidate. This is because accessing disk is detrimental to the performance.}

\ANNeX employs existing optimization techniques to reduce SSD access latency of reranking top-$R$ candidate NNs.


\outline{What is our approach? list out optimizations.}

\paragraph{\textbf{i) Batched, non-blocking multi-candidate reranking.}} 

Synchronous IO is slow due to its round trip delay. \ANNeX leverage asynchronous batching by combining $B$ candidate vectors in the candidate list into one big batch and submit $S = \lceil R/B \rceil$ asynchronous batched requests to SSD. Batched requests allow \ANNeX to exploit the internal parallelism of SSD. Asynchronous IO avoids blocking \ANNeX search and allows it to recompute a vector as soon as its full-view vector has been loaded into memory.
\ANNeX uses auto-tuning to identify the optimal combination of $B$ and $S$.



\later{This paper can become a focus on how to do post-verification effectively.
Is there a way to make the candidate list dynamic sized? Some pruning technique or heuristic would do.}

\outline{Direct IO, random access, not going to be used again. Not interfere with codebook.}
\later{
\begin{algorithm}[h!]
\caption{\hfill \ANNeX full-view reranking}
\label{algo:full-view-reranking}
  \begin{algorithmic}[1]
    \State \textbf{Input:} Query vector $q$, candidate short-list $R$, number of nearest neighbors $K$ to retrieve.
    	\label{lst:search:fullview-input}
    \State \textbf{Output:} Top-$K$ nearest neighbors.
    	\label{lst:search::output}
    \State \textbf{Hyperparameter:}  Batch size $B$, number of asynchronous submission $S$.
    	\label{lst:search:hyperparameter}
    \State \textbf{Initialization:} $TopK \leftarrow min\_priority\_queue(K)$ 
    	\label{lst:preview-step:hyperparameter}
    \ForAll {$s$ \textbf{in} 0..($S-1$)}
        \ForAll {$i$ \textbf{in} 0..($B-1$)}
            \label{lst:fullview-search:start}
            \State $req[i] \leftarrow prepare\_io\_requests(R[s \times B + i])$ 
            \label{lst:preview-search:prepare-io}
        \EndFor
        \State $handle \leftarrow submit\_async\_io(req)$
            \Comment{Non-blocking asynchronous batching submit.}
            \label{lst:preview-search:submit-io}
        \While {$reap\_cnt < B$}
            \State $num\_resp \leftarrow fetch\_io\_requests(resp)$ 
            \label{lst:fullview-search:fetch-io}
            \State $reap\_cnt += num\_resp$
            \ForAll {$i$ \textbf{in} 0..($num\_resp-1$)}
                \State $id, vec \leftarrow resp[i].id, resp[i].vec$ 
                \State $dist \leftarrow recompute\_distance(q, vec)$ 
                \State $TopK.add(dist, id)$ 
                \label{lst:fullview-search:reranking}
            \EndFor
            \label{lst:fullview-search:end}
        \EndWhile
    \EndFor
    \State return\ TopK
  \end{algorithmic}
\end{algorithm}
}

\paragraph{\textbf{ii) OS buffered I/O bypass.}}

\ANNeX employs \emph{direct IO} to bypass the system \emph{page cache} to suppress the memory interference caused by loading full-view vectors stored on SSD. 
Modern operating systems often maintain page cache, where the OS kernel stores page-sized chunks of files first into unused areas of memory, which acts as a cache.
In most cases, the introduction of page cache could achieve better performance. However, there are two main reasons we want \ANNeX to opt-out of system page cache: 1)  the reranking of candidates mostly incur random reads, which increases cache competition with a poor cache reuse characteristics; 2) page caching fullview vectors increases the memory cost to host \ANNeX index, decreasing its memory efficiency.

\outline{Give the post-validation algorithm or example. Including "hidden opportunity for asynchronous query submission. Also can add a naive loop based solution. Like how the DeepCPU paper explains WCS.}

\paragraph{Implementation.}
We implement the SSD-based reranking using the Linux NVMe Driver. The implementation uses the Linux kernel asynchronous IO (AIO) syscall interface, \code{io\_submit}, to submit IO requests and \code{io\_getevents} to fetch completions of loaded full-length vectors on SSD. Batched requests are submitted through an array of \code{iocb} struct at one. We open the file that stores the full-length vectors in \code{O\_DIRECT} mode to bypass kernel page cache.
Apart from Linux, most modern operating systems, including Windows, allow application to issue asynchronous, batched, and direct IO requests~\cite{linux-aio, windows-aio}.
There are also other advanced NVMe drivers like SPDK~\cite{spdk} and NVMeDirect~\cite{nvme-direct}, which offers even better performance on SSD by moving drivers into userspace and enable optimizations such as zero-copy accesses. We choose AIO for its simplicity and compatibility. 


\later{Dynamic Threshold for Candidate List Sizing?}

\later{Explain ComputePeak and Bandwidth. The goal is to come up an example that shows it is impossible to get a few millisecond latency if C1 is not large.\minjia{Not so well connected with the rest of the paper. Perhaps add it when having the equation to identify the execution time.}}

\section{Evaluation}
\label{sec:eval}

We evaluate \ANNeX and show how its design and algorithms contribute to its goals.




\subsection{Methodology}

\paragraph{\textbf{Workload.}}
We use \SIFT
and \Deep datasets for the experiments. 
\begin{itemize}
\item \emph{SIFT1M}~\cite{product-quantization} is a classical dataset to evaluate nearest neighbor search~\cite{sift}. It consists of one million  128-dimensional SIFT vectors, where each vector takes 512-byte to store~\footnote{http://corpus-texmex.irisa.fr/}. 
\item \emph{Deep10M} is a dataset that consists of 10 millions of 96-dimensional feature vectors generated by deep neural network~\cite{deep1B}~\footnote{http://sites.skoltech.ru/compvision/noimi/}. Each vector requires 384-byte memory. 
\end{itemize}


\begin{table*}[t]
	\newcommand{\colspc}{\hspace*{0.22em}}
    \small
    \renewcommand\sfsmaller{}
    \centering
\begin{tabular}{|@{\colspc}c@{\colspc}|c@{\colspc}|cc@{\colspc}@{\colspc}c@{\colspc}|cc@{\colspc}@{\colspc}c@{\colspc}|c@{\colspc}@{\colspc}c@{\colspc}@{\colspc}c@{\colspc}@{\colspc}c@{\colspc}|c@{\colspc}c@{\colspc}@{\colspc}c@{\colspc}@{\colspc}l@{\colspc}|}
\hline
	& \multirow{3}{*}{\textbf{$N_{scan}$}}                            & \multicolumn{6}{c|}{\textbf{IVFPQ}}                                                                      & \multicolumn{8}{c|}{\textbf{\ANNeX}}                                                                               \\ \cline{3-16}
\multirow{2}{*}{}        &        & \textbf{Recall}          & \textbf{Lat.}          & \textbf{Mem.}         & \textbf{Recall}          & \textbf{Lat.}          & \textbf{Mem.}         & \textbf{Recall}               & \textbf{Lat.}       & \textbf{Mem.}       & \textbf{VQ im.}           & \textbf{Recall}           & \textbf{Lat.}         & \textbf{Mem.}         & \textbf{VQ im.}      \\ \cline{3-16} 
                         &                             & \multicolumn{3}{c|}{$\approx $32 bytes per vector} & \multicolumn{3}{c|}{$\approx $64 bytes per vector} & \multicolumn{4}{c|}{$\approx $36 bytes per vector}                & \multicolumn{4}{c|}{$\approx $68 bytes per vector}           \\ \hline
\multirow{5}{*}{\textbf{SIFT1M}}  & \multirow{2}{*}{1024}                        & \multirow{2}{*}{0.679}                     & \multirow{2}{*}{8.8}  & \multirow{2}{*}{40}   & \multirow{2}{*}{0.850}           & \multirow{2}{*}{10.3} & \multirow{2}{*}{71}   & \textbf{0.894}                     & \multirow{2}{*}{0.5}  & \multirow{2}{*}{49}   & \multirow{2}{*}{\textbf{12.3X}}   & \textbf{0.898}                     & \multirow{2}{*}{0.9}  & \multirow{2}{*}{80}   & \multirow{2}{*}{\textbf{12.2X}}   \\  
						 &                         &                       &    &     &                       &   &     & $(N_{scan}:32)$                     &    &     &    & $(N_{scan}:32)$                     &    &    &     \\ \cline{2-16} 
  						 & 512        & 0.678                     & 5.4  & 40   & 0.849                     & 6.1                   & 71   & \textbf{0.989}            & 1.8  & 49   & \textbf{2.5X} & \textbf{0.999}            & 3.1  & 80   & \textbf{1.7X} \\ \cline{2-16} 
                         & 256        & 0.676                     & 3.2  & 40   & 0.847                     & 3.9                   & 71   & \textbf{0.986}            & 1.2  & 49   & \textbf{2.2X} & \textbf{0.995}            & 1.8  & 80   & \textbf{1.9X} \\ \cline{2-16} 
                         & 128        & 0.672                     & 2.5  & 40   & 0.840                     & 2.7                   & 71   & \textbf{0.976}            & 0.8  & 49   & \textbf{2.4X} & \textbf{0.984}            & 1.3  & 80   & \textbf{1.8X} \\ \cline{2-16} 
                         & 64         & 0.662                     & 2.0  & 40   & 0.820                     & 2.0                   & 71   & \textbf{0.948}            & 0.7  & 49   & \textbf{2.4X} & \textbf{0.955}            & 0.9  & 80   & \textbf{1.9X} \\ \hline \hline
                         &                             & \multicolumn{3}{c|}{$\approx $24 bytes per vector} & \multicolumn{3}{c|}{$\approx $48 bytes per vector} & \multicolumn{4}{c|}{$\approx $28 bytes per vector} & \multicolumn{4}{c|}{$\approx $52 bytes per vector} \\ \cline{2-16} 
\multirow{5}{*}{\textbf{Deep10M}} & \multirow{2}{*}{1024}                         & \multirow{2}{*}{0.602}                      & \multirow{2}{*}{18.5}  & \multirow{2}{*}{302}  & \multirow{2}{*}{0.866}                      & \multirow{2}{*}{24.6} & \multirow{2}{*}{531}  & \textbf{0.906}                     & \multirow{2}{*}{0.8} & \multirow{2}{*}{377}  & \multirow{2}{*}{\textbf{12.6X}}  & \textbf{0.921}                     & \multirow{2}{*}{1.4}  & \multirow{2}{*}{606}  & \multirow{2}{*}{\textbf{14.9X}}  \\ 
						&                          &                       &   &    &                       &   &    & $(N_{scan}:64)$                     &    &    &    & $(N_{scan}:64)$                     &    &    &    \\ \cline{2-16} 
						 & 512        & 0.601                     & 15.6 & 302  & 0.863                     & 18.9                  & 531  & \textbf{0.975}            & 5.4  & 377  & \textbf{2.3X} & \textbf{0.994}            & 4.6  & 606  & \textbf{3.6X} \\ \cline{2-16} 
                         & 256        & 0.599                     & 14.1 & 302  & 0.856                     & 15.7                  & 531  & \textbf{0.965}            & 3.4  & 377  & \textbf{3.3X} & \textbf{0.982}            & 2.8  & 606  & \textbf{4.9X} \\ \cline{2-16} 
                         & 128        & 0.594                     & 13.1 & 302  & 0.843                     & 14.2                  & 531  & \textbf{0.946}            & 2.4  & 377  & \textbf{4.3X} & \textbf{0.963}            & 1.9  & 606  & \textbf{6.4X} \\ \cline{2-16} 
                         & 64         & 0.580                     & 12.8 & 302  & 0.812                     & 13.1                  & 531  & \textbf{0.906}            & 2.0  & 377  & \textbf{5.2X} & \textbf{0.921}            & 1.4  & 606  & \textbf{8.0X} \\ \hline
\end{tabular}
\caption{Recall, latency (ms), memory (MB), VQ improvement of \ANNeX in comparison with \IVFPQ given different compression ratio and $N_{scan}$ for two datasets. 
}
\label{tbl:dq-improvement-ivfadc}
\end{table*}


\paragraph{\textbf{Evaluation metrics.}} Latency, memory cost, and recall are important metrics for ANNs. 
We measure query latency as the average elapsed time of per-query execution time in millisecond. The memory cost is calculated as the total allocated DRAM for the ANN index.
The recall of top-$K$ NNs is calculated as 
$\frac{|\xi(q) \bigcap \theta(q)|}{K}$, where $\xi(\cdot)$ and $\theta(\cdot)$ denote the set of exact NNs and the NNs given by the algorithm. 

In this paper, we also employ
\emph{VQ} (Vector--Query) as an important cost metric for ANN, inspired by the DQ (Document--Query) metric from web search engine~\cite{bitfunnel}. 
VQ is defined
as the product of the number of vectors per machine and the queries per second. 
For a given ANN workload with a total number of vectors $Y$ and total QPS of $Q$, an ANN solution requires $Y \times Q / VQ$ number of machines.  
The higher the VQ, the less machines and cost!  
VQ improvement over baseline ANNs is calculated as a product of the latency speedup and memory cost reduction rate.

\paragraph{Implementation.} \ANNeX is implemented in C++ on Faiss~\footnote{https://github.com/facebookresearch/faiss}, an open source library for approximate nearest neighbor search. The routing layer is implemented based on a C++ HNSW implementation from the HNSW authors~\footnote{https://github.com/nmslib/hnsw}. By default, Faiss evaluates latency with a very large batch size (10000) and report the execution time as the total execution time divided by the batch size. Such a configuration does not represent online serving scenarios, where requests often arrive one-by-one. We choose query batch size of 1 to represent a common case in online serving scenario. When batch size is small, Faiss has other performance limiting factors, \eg, unnecessary concurrency synchronization overhead,
which we address in \ANNeX. Also, neither Faiss nor HNSW parallelizes the execution of single query request. To make results consistent and comparable, \ANNeX does the same. 


\paragraph{\textbf{Experiment platform.}} We conduct the experiments on Intel Xeon Gold 6152 CPU (2.10GHz) with 64GB of memory and 1TB Samsung 960 Pro SSD. 
The server has one GPU (Nvidia GeForce GTX TITAN X) which is used for clustering during index construction.


\subsection{ANN Search Performance Comparison}
\label{subsec:eval-dq}

We first measure the performance results of \ANNeX and compare it with state-of-the-art ANN approaches: \IVFPQ~\footnote{IVFADC in the Faiss library.} and HNSW~\footnote{From the NMSLIB library: https://github.com/nmslib/nmslib}. Then we conduct a more detailed evaluation on the effect of different \ANNeX techniques.

\subsubsection{Comparison to \IVFPQ}

\begin{table*}[t]
  	\newcommand{\colspc}{\hspace*{0.95em}}
    \newcommand{\smallercolspc}{\hspace*{0.33em}}
    \small
    \renewcommand\sfsmaller{}
    \centering
\begin{tabular}{|l|c|@{\colspc}c@{\colspc}|@{\colspc}c@{\colspc}|@{\colspc}c@{\colspc}|c|@{\colspc}c@{\colspc}|@{\colspc}c@{\colspc}|@{\colspc}c@{\colspc}|c@{\colspc}|}
\hline
\multirow{2}{*}{}        & \multicolumn{4}{c|}{\textbf{HNSW}}                    & \multicolumn{4}{c|}{\textbf{\ANNeX}}     & \textbf{VQ}      \\ \cline{2-9}
                         & \textbf{Recall} & \textbf{efSearch} & \textbf{Latency} & \textbf{Memory} & \textbf{Recall} & $N_{scan}$ & \textbf{Latency} & \textbf{Memory} & \textbf{Improvement} \\ \hline
\multirow{4}{*}{\textbf{SIFT1M}}  & 0.993           & 1280              & 2.3                   & 588                & 0.991           & 1024                & 3.1                   & 49                 & \textbf{9.0X}        \\ \cline{2-10} 
                                  & 0.984           & 640               & 1.2                   & 588                & 0.989           & 512                 & 1.8                   & 49                 & \textbf{8.2X}        \\ \cline{2-10} 
                                  & 0.973           & 320               & 0.6                   & 588                & 0.976           & 128                 & 0.8                   & 49                 & \textbf{8.6X}        \\ \cline{2-10} 
                                  & 0.947           & 160               & 0.3                   & 588                & 0.948           & 64                  & 0.7                   & 49                 & \textbf{5.3X}        \\ \hline \hline
\multirow{4}{*}{\textbf{Deep10M}} & 0.998           & 1280              & 3.1                   & 4662               & 0.998           & 1024                & 6.7                   & 377                & \textbf{5.8X}        \\ \cline{2-10} 
                                  & 0.993           & 640               & 1.6                   & 4662               & 0.994           & 512                 & 3.9                   & 377                & \textbf{5.0X}        \\ \cline{2-10} 
                                  & 0.985           & 320               & 0.9                   & 4662               & 0.983           & 256                 & 2.6                   & 377                & \textbf{4.2X}        \\ \cline{2-10} 
                                  & 0.969           & 160               & 0.4                   & 4662               & 0.961           & 128                 & 2.0                   & 377                & \textbf{2.7X}        \\ \hline
\end{tabular}
\caption{Latency (ms), memory (MB), and VQ improvement of \ANNeX in comparison with HNSW under comparable recall target. }
\label{tbl:dq-improvement-hnsw}
\end{table*}

Table~\ref{tbl:dq-improvement-ivfadc} reports
the \emph{recall} (for $K=1$), latency, memory cost, and VQ improvement of \ANNeX, in comparison to \IVFPQ, for two datasets.
Both \IVFPQ and \ANNeX partition the input vectors into $N_{cluster}$ \cells (20K for \SIFT, and 200K for \Deep~\footnote{Training 10 million vectors on 200K clusters takes 5.8 hours on GPU.}) and generate the same PQ codebooks to encode all vectors. We vary the selected \cells $N_{scan}$ from 64 to 1024. This is the range we start to see that further increasing $N_{scan}$ leads to diminishing return of recall from \IVFPQ.
We also vary memory requirement by quantizing vectors with 
two different compression ratios (16X and 8X): $\approx$32, 64 bytes per vector for \SIFT, and $\approx$24, 48 bytes per vector for \Deep, by choosing different $M$.
Two main observations are in order.


\outline{1. Recall improvement. 2. Latency improvement with small sacrifice on memory. 3. VQ improvement. 4. VQ improvement given different Ms.}

First, the results show that \emph{\ANNeX provides significant improvement to the recall from about 0.58--0.68 to 0.90--0.99, while at the same time improving VQ by 1.7--8.0 times among tested configurations.} As expected, by varying $N_{scan}$ from 64 to 512, the recall and latency increase for both implementations as scanning more \cells increases both the likelihood and time of finding top-$K$ NNs. The recall of \IVFPQ is constantly lower than \ANNeX, and it starts to saturate at $N_{scan}=128$. In contrast, \ANNeX improves the recall by a large percentage. More importantly, it significantly improves the system's effectiveness by bringing the recall to the $0.90+$ range. This is contributed by \ANNeX's multi-view accuracy enhancement procedure. 
In terms of efficiency, \ANNeX overall speedups query latency in the range of 1.9--9.1 times, compared to \IVFPQ, with a slightly higher memory cost due to the additional memory cost of storing the routing layer and caching precomputation results. 
The \emph{VQ im.} column captures the VQ improvement where \ANNeX consistently outperforms \IVFPQ.

Second, \emph{\ANNeX improves VQ by 12.2--14.9 times to meet similar or higher recall target}. 
The highest recall \IVFPQ can get is still far below 1 
(\eg, at 0.679 when $N_{scan}=1024$ for \SIFT). This is not because \IVFPQ fails to scan \cells that include true NNs, but
because some true NNs cannot be differentiated by their PQ code based distance to the query even when they get scanned. 
In contrast, \ANNeX does not suffer as much from this issue and can achieve similar or higher recall by scanning a much smaller number of \cells (\eg, 32 and 64 \cells as marked by the parentheses below the recall measure), which also leads to a lower latency.

\subsubsection{Comparison to HNSW}
\label{subsec:eval-analysis}

We also compare \ANNeX with HNSW, the state-of-the-art NN proximity graph based approach. Since these two are very different solutions, we compare them by choosing configurations from both that achieve comparable recall targets (\eg, 0.95, 0.96, 0.97, 0.98, and 0.99)~\footnote{We build HNSW graph with $efConstruction$=200 and $OutD$=10. We trade-off accuracy and latency by varying $efSearch$ from 160 to 1280.}. 

As reported by Table~\ref{tbl:dq-improvement-hnsw}, {\it \ANNeX significantly and consistently outperforms HNSW, with an average VQ improvement of 3.5--9.0 times
among tested configurations.}
\ANNeX reduces the memory cost by around 12 times
compared to HNSW.
Although HNSW runs faster than \ANNeX in most cases, the latency gap between \ANNeX and HNSW decreases as we increase the recall target. This is presumably because for HNSW, further increasing $efSearch$ leads to little extra recall improvement but significantly more node exploration and distance computation when the recall is getting close to 1.

\subsection{Effect of Different Components}
\label{subsec:eval-analysis}

Next, we conduct an in-depth evaluation across different design points of \ANNeX.

\subsubsection{Latency of preview query processing}

\begin{figure}[H]
\centering
\small
\subfloat[][\SIFT cluster selection time \label{fig:in-memory-time-sift1m-cs}]{\includegraphics[width=0.53\linewidth]{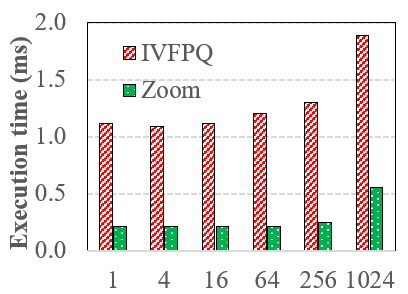}}
\subfloat[][\Deep cluster selection time \label{fig:in-memory-time-deep10m-cs}]{\includegraphics[width=0.47\linewidth]{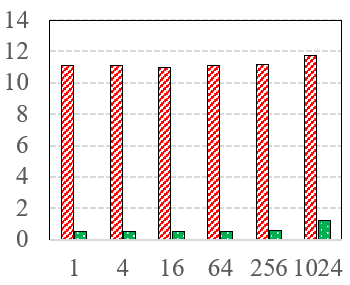}} \\ \vspace*{-1.2em}
\subfloat[][\SIFT vector scanning time \label{fig:in-memory-time-sift1m-vs}]{\includegraphics[width=0.53\linewidth]{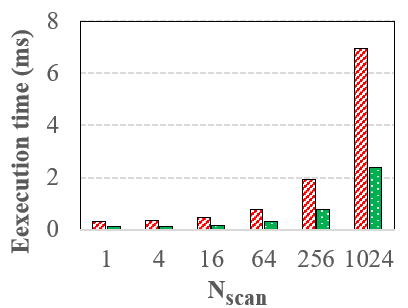}}
\subfloat[][\Deep vector scanning time \label{fig:in-memory-time-deep10m-vs}]{\includegraphics[width=0.47\linewidth]{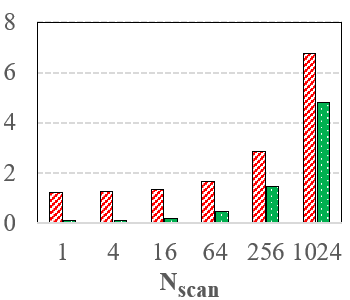}} \\ 
\caption{Effect of different components on the in-memory search query latency. The x-axis represents the number of selected and scanned \cells $N_{scan}$.}
\label{fig:in-memory-search-time}
\end{figure}

\later{TODO: Add performance results of different first-level cluster size.}
\later{TODO: Show some results on connectivity augmentation.}
Fig.~\ref{fig:in-memory-search-time} shows the breakdown of query latency on searching the in-memory preview index.

\paragraph{Latency of the routing layer.} Our results show that HNSW based routing yields significant improvements on \cell selection time compared with the exact search in \IVFPQ for both \SIFT (Fig.~\ref{fig:in-memory-time-sift1m-cs}) and \Deep (Fig.~\ref{fig:in-memory-time-deep10m-cs}). Overall, the HNSW routing layer speedups the cluster selection time by 3--6 times for \SIFT ($N_{cluster}=20K$) and 10--22 times for \Deep ($N_{cluster}=200K$). The HNSW routing layer offers higher speedup when the number of clusters $N_{cluster}$ is larger because the complexity of exact search is $O(D \times N_{cluster})$, 
whereas the complexity of HNSW based routing is $O(D \times \log{N_{cluster}})$, which is logarithmic to $N_{cluster}$. Therefore, the HNSW routing layer scales  better as $N_{cluster}$ increases and the improvement becomes more significant with larger $N_{cluster}$ sizes. 

\paragraph{Latency of PQ layer.} 

Fig.~\ref{fig:in-memory-time-sift1m-vs} and Fig.~\ref{fig:in-memory-time-deep10m-vs} illustrate the improvements of query latency of the PQ layer compared to \IVFPQ.
As $N_{scan}$ increases, the execution time of the PQ layer increases almost linearly for both \IVFPQ and \ANNeX. However, the PQ layer in \ANNeX consistently outperforms \IVFPQ by 2--13 times.
This is because \ANNeX optimizes the distance computation by reducing $2 \times M$ lookup-add operations
to $M + 1$ lookup-add per vector, significantly reducing memory accesses and mitigating memory bandwidth consumption.

\subsubsection{Accuracy of HNSW-based routing}

\begin{figure}[H]
\centering
\includegraphics[width=1\linewidth]{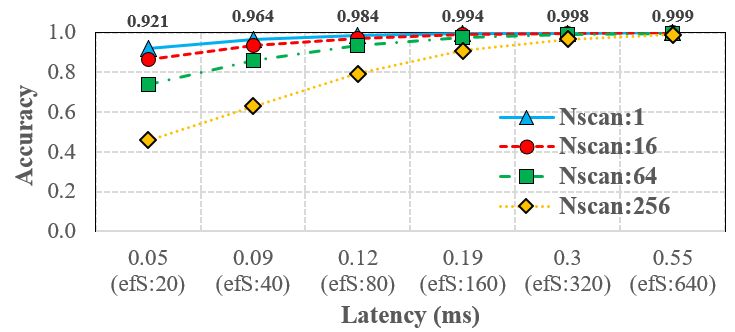}
\caption{Trade-offs between HNSW routing accuracy and latency on 200K centroids of \Deep. The y-axis represents the latency in millisecond. $(efS: X)$ in parentheses represents that the latency is obtained with $efSearch$ set to $X$.}
\label{fig:hnsw-routing-accuracy}
\end{figure}

Here we evaluate how accurately HNSW can identify $N_{scan}$ clusters compared with doing an exact search. The accuracy is the probability that $N_{scan}$ selected clusters are $N_{scan}$ true closest clusters to the query (essentially the same as recall). Fig.~\ref{fig:hnsw-routing-accuracy} reports accuracy of searching $200K$ clusters of the \Deep dataset when $N_{scan}$ is 1, 16, 64, and 256, which correspond to performing $K$-NN search by HNSW with $K$ equals to 1, 16, 64, and 256 respectively.
Overall, gradually increasing $efSearch$ leads to higher accuracy at the expense of increased routing latency. HNSW-based routing can achieve fairly high accuracy (\eg. when $efSearch$ is 320) for various $N_{scan}$. Further increasing $efSearch$ (\eg, to 640) leads to little extra accuracy improvement but significantly longer latency because the accuracy is getting close to 1. 
We also observe that under the same $efSearch$, larger $N_{scan}$ sometimes leads to slightly worse accuracy if $efSearch$ is not big enough (\eg, $efSearch$ is less than 160), as the closest clusters not visited during the routing phase are definitely lost. In our experiments, we choose a sufficiently large $efSearh$ (\eg, 320) to get close to unity accuracy for the routing layer. 

\subsubsection{Sensitivity of $K$ and $R$}
\label{subsub:sensitivity-k-and-r}

In practice, different applications might require to retrieve different $K$ NNs. 
Table~\ref{tbl:k-and-r-sensitivity} shows the recall of \ANNeX at different $K$ and $R$ compared to \IVFPQ varying $N_{scan}$ from 1 to 1024 on the \Deep dataset. We make two observations.  First, \ANNeX offers significant recall improvement for K=1 as well as K > 1. In both cases, the recall of \IVFPQ has reached its plateau around 0.60 and 0.70, whereas \ANNeX consistently brings the recall to $0.98+$. Second, a small $R$ can sharply improve the recall. Although larger $R$ is better for getting higher recall, we observe that further increasing $R$ from 10 to 100 when $K=1$ or from 50 to 100 when $K=10$ does not bring a lot more improvement on recall, indicating that a small $R$ is often sufficient to get high recall.

\begin{table}[H]
	\newcommand{\colspc}{\hspace*{0.21em}}
    \small
    \renewcommand\sfsmaller{}
    \centering
\begin{tabular}{|@{\colspc}r@{\colspc}|l|ll|l|ll@{\colspc}|}
\hline
\multirow{3}{*}{\textbf{$N_{scan}$}} & \multicolumn{3}{c|}{\textbf{K=1}}                                            & \multicolumn{3}{c|}{\textbf{K=10}}                                           \\ \cline{2-7} 
                  & \multirow{2}{*}{\textbf{IVFPQ}} & \multicolumn{2}{c|}{\textbf{\ANNeX}} & \multirow{2}{*}{\textbf{IVFPQ}} & \multicolumn{2}{c|}{\textbf{\ANNeX}} \\ \cline{3-4} \cline{6-7} 
                  &                        & R=10                 & R=100               &                        & R=50                 & R=100               \\ \hline
1                 & 0.245                  & 0.291                & 0.292               & 0.194                  & 0.200                & 0.200               \\ \hline
4                 & 0.396                  & 0.518                & 0.521               & 0.382                  & 0.414                & 0.414               \\ \hline
16                & 0.519                  & 0.756                & 0.763               & 0.559                  & 0.667                & 0.669               \\ \hline
64                & 0.580                  & 0.906                & 0.920               & 0.657                  & 0.864                & 0.868               \\ \hline
256               & 0.599                  & 0.965                & 0.983               & 0.691                  & 0.958                & 0.966               \\ \hline
1024              & 0.602                  & 0.978                & 0.998               & 0.697                  & 0.983                & 0.994               \\ \hline
\end{tabular}
\caption{Effect of $K$ and $R$ on recall for \IVFPQ and \ANNeX.}
\label{tbl:k-and-r-sensitivity}
\end{table}



\subsubsection{Performance of SSD-assisted reranking}

We measure the reranking latency
at the full-view step.
Without optimizations, it takes 68us to rerank a single candidate vector using  synchronous IO without batching. This is slow and it would take a few milliseconds to rerank 100 candidates. 

\begin{figure}[H]
\centering
\includegraphics[width=1\linewidth]{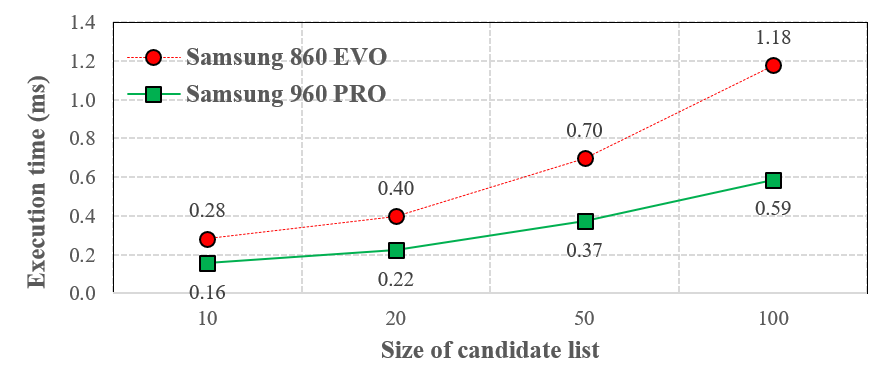}
\caption{Latency of reranking on two SSDs varying $R$.}
\label{fig:SSD-based-reranking}
\end{figure}

Fig.~\ref{fig:SSD-based-reranking} shows the latency impact of the batched, non-blocking multi-candidate reranking employed by \ANNeX, varying the size of candidate list $R$ from 10 to 100. For $R$ in this range, we have found that a simple strategy of setting batch size $B$ to $R$ and asynchronous submission count $S$ to 1 already performs much better than synchronous, non-batched reranking. As the candidate list size increases, the reranking time increases almost linearly. With Samsung Pro 960, \ANNeX can rerank up to 100 candidates of vector length 512-byte in less than 0.6ms. As \ANNeX requires only a small set of candidates to be reranked, 0.6ms can already cover a good range of $R$.

We also compare the reranking latency with another SSD Samsung 860 EVO (1TB), to evaluate the sensitivity of the latency on different SSDs. The conclusion still holds for this consumer-grade SSD. The reranking latency roughly get doubled, but it can still rerank 100 candidates in less than 1.2ms.
Overall, the results indicate that \ANNeX can leverage SSDs for the full-view reranking with a small increase on overall latency.

\section{Related Work}
\label{sec: related}

\paragraph{Tree-based ANN.}

One large category of ANN algorithm is tree-based ANN, such as KD-tree~\cite{kd-tree} and VP-tree~\cite{vp-tree}. These approaches work well in low dimensions.
However, 
the complexity of these approaches is $O(D \times N^{1-1/D})$, which 
is not more efficient than a brute-force distance computation at high dimension~\cite{worst-case-kdtree}. 

\paragraph{Other compact code based approaches.}
Another large body of existing ANN work relies on hashing~\cite{lsh, polysemous-code}, which approximates the similarity between two vectors using hashed codes or Hamming codes.
One of the well-known representatives is Locality-Sensitive Hashing (LSH). LSH has the sound probabilistic theoretical guarantees on the query result quality, but 
product quantization and its extensions combined with inverted file index
have been proven to be more effective on large-scale datasets than hashing-based approaches~\cite{ann-experiments-analysis,lopq}.


\paragraph{Hardware accelerators.} Apart from CPU, researchers and practitioners are also looking into using GPU for vector search~\cite{billion-scale-search-on-gpus,rbe-gpu-exhaustive-nn-search,ann-on-gpu}. However, GPUs also face the same effectiveness and efficiency challenges as their memory is even more limited than CPUs.  Although GPUs offer high throughput for offline ANN search, 
small batch size during online serving can hardly make full usage of massive GPU cores either, rendering low GPU efficiency.
Furthermore, people propose to use specialized hardware such as FPGA~\cite{ann-on-fpga} to serve ANN, but it often requires expert hardware designers and long development cycles to obtain high performance.




\section{Conclusion}
\label{sec:conclusion}

Vector search becomes instrumental with the major advances in deep learning based feature vector extraction techniques. It is used by many information retrieval services, and it is crucial to conduct search with high accuracy, low latency, and low memory cost. We present \ANNeX, an ANN solution that employs a multi-view approach and takes the full storage architecture into consideration that greatly enhance the effectiveness and efficiency of ANN search. \ANNeX uses SSD to maintain full-view vectors and performs reranking on a list of candidates selected by a preview step as an accuracy enhancement procedure. It also improves the overall system efficiency through efficient routing and optimized distance computation. 
\ANNeX achieves an order of magnitude improvements on efficiency while attaining equal or higher accuracy, compared with the-state-of-the-art. 
We conclude that \ANNeX is a promising approach to ANN. We hope that \ANNeX will enable future system optimization works on vector search in high dimensional space. 
\later{
Our next steps include expanding \ANNeX to even larger scale datasets and exploring more efficient use of SSDs for ANN serving.
}


\minjia{\sout{TODO: $C_{pq}$ vs. $C_2$}}
\minjia{\sout{TTODO: $C_{coarse}$ vs. $C_1$}}
\minjia{TODO: Examine and fix bib}
\minjia{\sout{TODO: speedup X vs. $\times$}}

\minjia{\sout{TODO: Shorten SSD part.}}
\minjia{\sout{TODO: Shorten distance computation.}}
\minjia{\sout{TODO: Adjust graphs.}}
\minjia{\sout{TODO: Add in Fig.2 enlarged results. Like Focus.}}
\minjia{\sout{TODO: Make references in order.}}
\minjia{\sout{TODO: query should be $q$, data points use $y$.}}
\minjia{TODO: memory efficiency, memory cost, memory overhead.}
\minjia{TODO: X vs. times consistency. Use the later one.}
\minjia{TODO: Recall \% vs. 0.99.}
\minjia{TODO: Avoid using short-list because R is larger than K.}
\minjia{Preview vector layer vs. pq layer. vs. pq index}
\minjia{Preview preview vector vs. pq vector}
\minjia{TODO: Revise eval based on "effectiveness and efficiency.}

\medskip

\clearpage

\bibliographystyle{plain}
\bibliography{main-asplos19}

\end{document}